\documentclass[10pt,twocolumn,letterpaper]{article}

\usepackage[pagenumbers]{cvpr}

\usepackage{cuted} 

\usepackage{currfile} 

\usepackage{caption}

\usepackage{booktabs}
\usepackage{multirow}
\usepackage{amsmath,amssymb}
\usepackage{xspace}
\usepackage{algorithm}
\usepackage{algpseudocode}
\usepackage{tikz}
\usetikzlibrary{arrows.meta,positioning,fit,backgrounds,calc}

\newcommand{\method}{SHINE\xspace}

\newcommand{\best}[1]{\textbf{#1}}
\newcommand{\cA}{\mathcal{A}}
\newcommand{\cO}{\mathcal{O}}
\newcommand{\cQ}{\mathcal{Q}}

\newcommand{\cL}{\mathcal{L}}

\definecolor{cvprblue}{rgb}{0.21,0.49,0.74}
\usepackage[breaklinks,colorlinks,allcolors=cvprblue]{hyperref}

\def\confName{}
\def\confYear{}

\title{Searching with Sparse Histories: Cross-Task Architecture Search\\from Incomplete Search Logs and Zero-Cost Proxies}

\author{
Yuefeng Xu\\
Graduate School of Science and Technology\\
Niigata University, Niigata, Japan\\
{\tt\small f25l005c@mail.cc.niigata-u.ac.jp}
\and
Jun Yu\thanks{Corresponding author.}\\
Institute of Science and Technology\\
Niigata University, Niigata, Japan\\
{\tt\small yujun@ie.niigata-u.ac.jp}
}

\begin{document}
\maketitle
\begin{abstract}
Neural architecture search (NAS) on a new task is expensive because every candidate must be trained to be evaluated.
Cross-task NAS reduces this cost by reusing knowledge from previously searched tasks, but existing methods assume either dense performance tables on the source tasks, a single source task, or no source evaluations at all.
In practice, a team that has searched before holds \emph{sparse search histories}: several past tasks, each with a small and different subset of evaluated architectures.
We study cross-task NAS from such incomplete histories.
We first show that standard transfer heuristics break down in this regime, and that the two cheap sources of prior knowledge available to a practitioner---sparse historical logs and zero-cost proxies---fail on \emph{different} targets: proxies on dense-prediction tasks, sparse logs on tasks that are negatively correlated with the sources.
We then propose \method, a risk-robust transfer strategy that (i) starts the search from a portfolio covering the champions of each imputed source task and (ii) pretrains a pairwise ranker jointly on the sparse source tasks and on zero-cost proxies treated as pseudo-tasks, before adapting it to the target inside a weak-predictor search.
\method is not uniformly better on average than a strong proxy-only ranker; its value lies in substantially mitigating the severe failures of either knowledge source while retaining competitive average performance.
On TransNAS-Bench-101 with 50 target evaluations and 50 evaluated architectures per source, it mitigates the worst failure of each source (Micro autoencoding: worst-10\% rank 290 with proxies alone \vs 43; Macro room layout: 465 with logs alone \vs 45), and matches published methods in typical performance (rank of mean 2.00 on Macro \vs 2.25 for TRP and, on the four metric-compatible targets, better than Arch-Graph on both spaces).
Over 13 cell-space targets, including seven trained from scratch on new datasets and four targets run only after the analysis plan was fixed, \method improves over the proxy-only ranker on 10 of 13 targets in both mean and worst-case performance (task-level $p{\le}0.016$; 4 of 4 on the new targets) and is on par with source logs alone, which, however, fail severely on the macro space.
\end{abstract}

\section{Introduction}
\label{sec:intro}

Neural architecture search (NAS) has produced competitive networks for classification, dense prediction and many other vision problems~\cite{zoph2017nas,real2019rea,liu2019darts,elsken2019survey}.
Its central obstacle remains cost: the quality of an architecture is only known after it has been trained, so each search spends most of its budget on evaluations.
The cost is paid again for every new task, even when the search space and the training protocol are unchanged, because the ranking of architectures shifts from task to task~\cite{duan2021transnas,zamir2018taskonomy}.

\emph{Cross-task} NAS addresses this by transferring knowledge from previously searched tasks~\cite{chen2020catch,lee2021metad2a,huang2022archgraph,li2025trp}.
Existing approaches differ in the kind of source knowledge they assume.
Benchmark-driven analyses and direct-transfer baselines assume \emph{dense} source tables, \ie every architecture has been evaluated on every source task~\cite{duan2021transnas}.
Arch-Graph~\cite{huang2022archgraph} pretrains a relation predictor on a \emph{single} source task.
TRP~\cite{li2025trp} avoids source evaluations altogether and pretrains on \emph{zero-cost proxies}~\cite{abdelfattah2021zerocost,mellor2021nwot}.
None of these matches the situation of a practitioner who has searched before: such a team owns \emph{sparse search histories}---logs from several past tasks, each covering only the few dozen to few hundred architectures that happened to be trained on it, with little overlap between tasks.

This paper asks how to best use such incomplete histories when searching on a new task under a small evaluation budget.
We find that the question is not answered by existing heuristics.
Aggregating source rankings into a consensus prior, a common default, performs poorly because high-performing architectures are task specific.
Covering the champions of each source task is a far stronger strategy when source tables are dense, but its advantage collapses when each source is evaluated on only $m{=}50$ architectures (\cref{sec:why}).
Zero-cost proxies are a natural complement, as they are available for every architecture at negligible cost, yet we show that they are unreliable precisely on the dense-prediction tasks where transfer matters most (\cref{sec:complementarity}).

We propose \method (\textbf{S}parse-\textbf{H}istory \textbf{IN}formed s\textbf{E}arch), a simple method that unifies the two cheap sources of knowledge.
\method treats every sparse source task and every zero-cost proxy as a ranking task over the same architecture space and pretrains a pairwise ranker on all of them jointly.
The search starts from a portfolio that covers the champions of each source task after imputation, and it adapts the ranker on the target inside a weak-predictor search loop that ends with an exploitation phase.
\method requires neither dense source tables nor task embeddings, and its overhead is negligible compared with a single architecture evaluation.
We position \method as a \emph{risk-robust} transfer strategy rather than a uniformly superior one: in typical runs it is on par with the better single knowledge source, and its benefit is that it substantially mitigates the severe failures that each source suffers on some target.

Our contributions are:
\begin{itemize}
\item We formulate cross-task NAS from \emph{sparse search histories}, a setting that reflects how search knowledge accumulates in practice, and show that consensus priors and source-portfolio heuristics fail in it.
\item We propose \method, which combines a source-champion portfolio with multi-task pairwise ranker pretraining over sparse source tasks and zero-cost pseudo-tasks.
\item We show that historical logs and zero-cost proxies fail on different targets, and that combining them mitigates the severe failures of either, reduces tail risk relative to proxies, while typical-case performance stays on par with the better single source.
\item On TransNAS-Bench-101~\cite{duan2021transnas}, \method matches published methods in typical performance (better than Arch-Graph~\cite{huang2022archgraph} on the metric-compatible targets, on par with TRP~\cite{li2025trp} on Macro); in a pooled analysis over 13 targets, seven of them trained from scratch on new datasets, with the analysis plan fixed before the last four were run, \method significantly improves over the proxy-only ranker at the task level in both mean and worst-case performance; MedMNIST-NAS-Bench~\cite{medmnistnasbench} further validates the historical-log component.
\end{itemize}

\section{Related Work}
\label{sec:related}

\paragraph{Predictor-based NAS.}
Performance predictors reduce the number of trained architectures by ranking candidates before evaluation~\cite{white2021bananas,wu2021weaknas,white2023insights}.
BANANAS~\cite{white2021bananas} showed that path-based encodings substantially improve predictors on cell search spaces; weakNAS~\cite{wu2021weaknas} iteratively fits weak predictors and progressively narrows the sampling region; and simple local search is a strong baseline on tabular benchmarks~\cite{white2021localsearch}.
Pairwise and listwise ranking losses~\cite{burges2005ranknet} are preferred over regression because search only needs the order of architectures.
\method uses the same search template but focuses on what the predictor is pretrained on.

\paragraph{Cross-task and transferable NAS.}
TransNAS-Bench-101~\cite{duan2021transnas} provides seven vision tasks over two shared search spaces and showed that direct architecture transfer is often worse than random search.
CATCH~\cite{chen2020catch} meta-learns a controller with task context, MetaD2A~\cite{lee2021metad2a} generates architectures from dataset embeddings, and EMT-NAS~\cite{liao2023emtnas} and KTNAS~\cite{ktnas2025} exchange knowledge between tasks searched simultaneously.
Arch-Graph~\cite{huang2022archgraph} pretrains a pairwise relation predictor on one source task with Task2Vec embeddings~\cite{achille2019task2vec} and fine-tunes it on 30 target evaluations.
TRP~\cite{li2025trp} pretrains a relativistic predictor on zero-cost proxy votes and fine-tunes it on a few dozen target evaluations.
Multi-Predict~\cite{akhauri2023multipredict} and HELP~\cite{lee2021help} transfer predictors across tasks or devices with few samples.
Unlike these works, we assume several \emph{sparsely} evaluated source tasks and show how to combine them with proxies.

\paragraph{Zero-cost proxies.}
Zero-cost proxies score untrained networks from a single minibatch~\cite{abdelfattah2021zerocost,mellor2021nwot,tanaka2020synflow,lin2021zennas}.
NAS-Bench-Suite-Zero~\cite{krishnakumar2022nbsz} found that their reliability varies strongly across search spaces and tasks.
Proxies have also been used as input features of single-task predictors~\cite{white2021powerful}.
We confirm their unreliability on dense-prediction tasks and use them as auxiliary pretraining tasks alongside sparse source logs rather than as a ranking.

\paragraph{Warm-starting from previous tasks.}
In hyperparameter optimisation, configurations that performed well on previous datasets are used to initialise the search~\cite{feurer2015initializing,wistuba2015sequential}, often as a greedy portfolio~\cite{pfisterer2021defaults,feurer2022autosklearn2}, and source surrogates can be ensembled with ranking-based weights~\cite{feurer2018rgpe}.
\method adopts the portfolio idea for initialisation and extends it to sparse sources by imputation.

\section{Cross-Task NAS from Sparse Search Histories}
\label{sec:setting}

\paragraph{Setting.}
Let $\cA$ be a finite search space of $N$ architectures and let $y_j(a)$ denote the quality of architecture $a\in\cA$ on task $j$, oriented so that larger is better.
A \emph{search history} consists of $S$ source tasks.
For source $s$ we observe only a subset $\cO_s\subset\cA$ with $|\cO_s|=m\ll N$ and the corresponding values $\{y_s(a)\}_{a\in\cO_s}$; the subsets of different sources are drawn independently.
In addition, $K$ zero-cost proxies $z_1,\dots,z_K$ can be computed for every architecture at negligible cost.
On a new target task $t$, an algorithm sequentially selects architectures to train; after $B$ evaluations it returns the best evaluated architecture.
No information about $y_t$ is available except through these $B$ evaluations.

\paragraph{Evaluation.}
Following Arch-Graph~\cite{huang2022archgraph}, we report the global rank of the returned architecture within $\cA$ (1 is best), averaged over target tasks and random seeds.
For comparability with published tables, which report the metric value of the returned architecture averaged over trials, we also report the global rank of that average value.
Each target is held out in turn and the remaining tasks act as sources (leave-one-task-out).

\subsection{Why naive transfer fails with sparse histories}
\label{sec:why}

Before introducing our method we examine two natural heuristics on TransNAS-Bench-101 (setup in \cref{sec:exp}).
\emph{Consensus} ranks architectures by their mean percentile rank over sources.
\emph{Champion portfolio} greedily selects architectures to minimise the summed regret over source tasks~\cite{wistuba2015sequential,pfisterer2021defaults}, which effectively visits the best architecture of each source first.
With dense source tables, the champion portfolio is remarkably strong: after 50 target evaluations its average rank is 11.0 on Macro and 15.2 on Micro, close to Arch-Graph (5.2 and 12.2), whereas consensus reaches only 90.5 and 44.2.
Top architectures are thus task specific, and averaging hides them.
However, when each source reveals only $m{=}50$ random architectures, the portfolio degrades to an average rank of 24.3 (Macro) and 44.3 (Micro), and a rule that selects architectures with high cross-source disagreement collapses entirely (122.6 and 516.2).
A method for sparse histories must therefore generalise beyond the few observed source architectures and must not rely on the sources alone.

\section{Method}
\label{sec:method}

\begin{figure}[t]
\centering
\includegraphics[width=\linewidth]{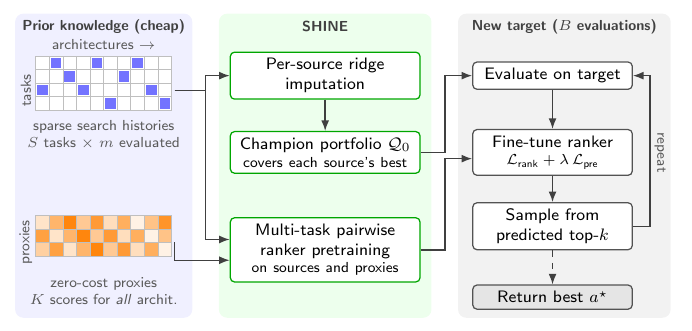}
\caption{\textbf{Overview of \method.} Sparse source logs (each past task has only $m$ evaluated architectures) are imputed per source to build a champion portfolio $\mathcal{Q}_0$ that initialises the search. A pairwise ranker is pretrained jointly on all sparse source tasks and on zero-cost proxies treated as pseudo-tasks; on the new target it is repeatedly fine-tuned on the evaluated architectures and used to sample the next ones from its predicted top-$k$ until the budget $B$ is spent.}
\label{fig:overview}
\end{figure}

\method has three components (\cref{fig:overview}): a champion portfolio built from imputed source rankings (\cref{sec:portfolio}), a pairwise ranker pretrained on sparse sources and zero-cost pseudo-tasks (\cref{sec:pretrain}), and a target-side search that adapts this ranker (\cref{sec:search}).

\subsection{Champion portfolio from imputed sources}
\label{sec:portfolio}
Each architecture is described by a structural encoding $x_a$ (\cref{sec:exp}).
For every source $s$ we fit a ridge regressor with pairwise feature interactions to the percentile ranks observed on $\cO_s$ and predict a complete ranking $\hat r_s(a)$; observed architectures keep their true order.
The initial target queries $\cQ_0$, $|\cQ_0|=K_0$, greedily minimise the summed normalised regret over sources,
\begin{equation}
\cQ_0=\arg\min_{|\cQ|=K_0}\sum_{s=1}^{S}\Big[\max_{a\in\cA}\hat r_s(a)-\max_{a\in\cQ}\hat r_s(a)\Big],
\label{eq:portfolio}
\end{equation}
with ties broken by the mean imputed rank.
The first $S$ greedy steps visit the (imputed) champion of each source, after which the portfolio refines the source-specific top regions.
Unlike a consensus prior, the portfolio deliberately covers architectures that are excellent on \emph{some} source, which is where target champions are most often found (\cref{sec:why}).

\subsection{Multi-task pairwise ranker pretraining}
\label{sec:pretrain}
We use a shared encoder $f_\theta$ (a two-layer MLP) with one linear head $(w_j,b_j)$ per task, $g_j(a)=w_j^\top f_\theta(x_a)+b_j$.
The pretraining tasks $\mathcal{J}$ comprise the $S$ sparse sources, each supervised only on $\cO_s$, and the $K$ zero-cost proxies, each supervised on all of $\cA$ by its own ranking.
Treating proxies as \emph{pseudo-tasks} lets the encoder learn architecture regularities from dense but noisy signals while the sparse sources contribute accurate but incomplete task-specific rankings.
All tasks are trained with a pairwise logistic (RankNet) loss~\cite{burges2005ranknet},
\begin{equation}
\cL_{\text{pre}}=\frac{1}{|\mathcal{J}|}\sum_{j\in\mathcal{J}}\mathbb{E}_{(a,b)\sim\mathcal{P}_j}\,\ell\big(\delta_j(a,b),\,g_j(a)-g_j(b)\big),
\label{eq:pretrain}
\end{equation}
where $\ell(\delta,u)=\log(1+e^{-\delta u})$, $\delta_j(a,b)=\operatorname{sign}(y_j(a)-y_j(b))$, and $\mathcal{P}_j$ samples pairs among the architectures labelled for task $j$.
Because only rankings are used, scales of different metrics and proxies need no calibration.
We train an ensemble of $E{=}4$ members and initialise the target head of each member as the mean of its pretraining heads.
The proxy values are also appended to $x_a$ as input features.

\subsection{Target adaptation and search}
\label{sec:search}
After observing the target set $\cQ$, each member is fine-tuned from its pretrained state on target pairs with the pretraining loss as a regulariser,
\begin{equation}
\cL_{\text{tgt}}=\cL_{\text{rank}}\big(g_t;\{(a,y_t(a))\}_{a\in\cQ}\big)+\lambda\,\cL_{\text{pre}},
\label{eq:target}
\end{equation}
and the ensemble-averaged standardised scores rank the unobserved architectures.
Following the weak-predictor paradigm~\cite{wu2021weaknas}, we alternate between refitting and querying: each round samples a batch of $b$ architectures from the predicted top-$k$ region, with $k$ shrinking geometrically, and the last $F$ evaluations exploit the predicted best architectures.
The regulariser keeps the ranker close to transferable structure when few target labels are available, while the target head absorbs task-specific deviations.
\cref{alg:method} summarises the procedure.

\begin{algorithm}[t]
\caption{\method}
\label{alg:method}
\small
\begin{algorithmic}[1]
\Require sparse logs $\{(\cO_s,y_s)\}_{s=1}^S$, proxies $\{z_k\}_{k=1}^K$, budget $B$
\State $\hat r_s\gets$ ridge imputation of each source; $\cQ\gets\cQ_0$ via \cref{eq:portfolio}; evaluate $\cQ$
\State pretrain ensemble $\{g^{(e)}\}_{e=1}^{E}$ with \cref{eq:pretrain}
\For{$\ell=1,2,\dots$ \textbf{while} $|\cQ|<B-F$}
  \State fine-tune each $g^{(e)}$ with \cref{eq:target}; $p\gets\frac1E\sum_e\mathrm{std}(g^{(e)}_t)$
  \State sample $b$ unseen architectures from top-$k_\ell$ of $p$; evaluate; add to $\cQ$
\EndFor
\While{$|\cQ|<B$}
  \State fine-tune; add the top-$F/2$ unseen architectures of $p$ to $\cQ$; evaluate
\EndWhile
\State \Return $\arg\max_{a\in\cQ}y_t(a)$
\end{algorithmic}
\end{algorithm}

\paragraph{Implementation.}
We use $B{=}50$, $K_0{=}0.2B$, $F{=}0.4B$, $b{=}5$, $k$ decreasing from 400 to 20, hidden width 128, 400 pretraining steps, 120 fine-tuning steps per round and $\lambda{=}0.3$.
All hyper-parameters are shared across search spaces and targets.
The full search, including all pretraining and fine-tuning, takes about one GPU-minute, negligible compared with a single architecture evaluation.

\section{Experiments}
\label{sec:exp}

\subsection{Setup}
\paragraph{Benchmarks.}
TransNAS-Bench-101~\cite{duan2021transnas} contains seven tasks derived from Taskonomy~\cite{zamir2018taskonomy}---object and scene classification, room layout, jigsaw, semantic segmentation, surface normals and autoencoding---evaluated on a cell-based Micro space (4{,}096 architectures) and a Macro skeleton space (3{,}256 architectures).
Following Arch-Graph~\cite{huang2022archgraph}, jigsaw is used only as a source and the remaining six tasks are targets; for each target all other tasks are sources.
For independent validation we use the public MedMNIST-NAS-Bench~\cite{medmnistnasbench}, a tabular benchmark on four MedMNIST v2 classification tasks~\cite{yang2023medmnist} (BreastMNIST, PneumoniaMNIST, DermaMNIST, BloodMNIST) in which every architecture is trained from scratch for 100 epochs with three seeds.
Its Tiny space contains 768 architectures from three macro families and its Small space 5{,}016 architectures with additional templates, attention and down-sampling choices.
Search observes validation AUROC, and we report the test AUROC of the architecture selected by validation.

\paragraph{Protocol.}
Unless stated otherwise, the target budget is $B{=}50$ evaluations, the same as Arch-Graph, and each source task reveals $m{=}50$ randomly chosen architectures.
Zero-cost proxies are the twelve proxies of NAS-Bench-Suite-Zero~\cite{krishnakumar2022nbsz} computed on the object-classification task, as in TRP~\cite{li2025trp}.
Five of them (fisher, grad-norm, grasp, snip, plain) use the labels of one minibatch of 32 images; to avoid leaking target labels, we use proxies computed on scene classification when object classification is itself the target, and report the original setting as a control in \cref{sec:main}. (All Taskonomy tasks share the same images.) In the real-training experiments (\cref{sec:real}) the proxies never see target data.
Structural encodings are BANANAS-style path encodings for the Micro cell~\cite{white2021bananas} and per-layer down-sampling and widening indicators for Macro.
Results are averaged over 20 seeds that resample the source logs and the search randomness (10 seeds for the source-density sweep and the budget sweep).
\method was developed on TransNAS-Bench-101 and all hyper-parameters were frozen before the final runs; MedMNIST-NAS-Bench was not used during development.
Code, source-log samples and all real-training records will be released.
The 20 seeds of a target share the same task, so seed-level tests only capture robustness to search randomness, not generalisation to new tasks.
For TransNAS-Bench-101 comparisons we therefore report the number of targets (out of six) on which a method is better after averaging over seeds, a task-level paired Wilcoxon test (whose smallest attainable two-sided $p$-value with six targets is 0.031) and, where useful, a 95\% hierarchical bootstrap interval of the mean difference (resampling targets, then seeds); ablation $p$-values are Holm-corrected.
We also report the mean of the worst 10\% of runs (\emph{worst-10\%}) as a robustness measure.

\paragraph{Baselines.}
We compare with the published results of single-task NAS (random search, REA~\cite{real2019rea}, weakNAS~\cite{wu2021weaknas}), transferable NAS (CATCH~\cite{chen2020catch}, transfer variants of weakNAS, Arch-Graph~\cite{huang2022archgraph}) and TRP~\cite{li2025trp}.
To control for implementation details we also run, under our exact protocol, a ZC-only ranker pretrained on proxy pseudo-tasks without any source logs (TRP-style), a sources-only ranker without proxies, and tree-based and ensemble variants.

\subsection{Main results}
\label{sec:main}

\begin{table*}[t]
\centering
\small
\caption{\textbf{Cross-task search on TransNAS-Bench-101 with 50 target evaluations.} Global rank of the returned architecture (lower is better). ``6 tasks'': mean rank over all six targets. ``4 tasks'': rank of the mean metric value over the four targets whose metric matches published tables (object/scene classification, autoencoding, normals), the convention of~\cite{huang2022archgraph,li2025trp}. ``Worst 10\%'': mean rank of the worst 10\% of runs. Published numbers are taken from~\cite{huang2022archgraph}; TRP numbers are converted from the metric values reported in~\cite{li2025trp}. Our implementation: $m{=}50$ evaluated architectures per source, 20 seeds, leakage-free proxies.}
\label{tab:main}
\begin{tabular}{llcccccc}
\toprule
& & \multicolumn{2}{c}{6 tasks (mean rank)} & \multicolumn{2}{c}{4 tasks (rank of mean)} & \multicolumn{2}{c}{Worst 10\%}\\
\cmidrule(lr){3-4}\cmidrule(lr){5-6}\cmidrule(lr){7-8}
Method & Source knowledge & Macro & Micro & Macro & Micro & Macro & Micro\\
\midrule
Random search            & none                 & 59.26 & 85.61 & -- & -- & -- & -- \\
REA~\cite{real2019rea}    & none                 & 41.03 & 38.50 & -- & -- & -- & -- \\
weakNAS~\cite{wu2021weaknas} & none              & 10.49 & 20.03 & -- & -- & -- & -- \\
Arch-Graph-single         & none                 & 12.68 & 22.15 & -- & -- & -- & -- \\
\midrule
CATCH~\cite{chen2020catch} & jigsaw + others      & 37.72 & 63.49 & -- & -- & -- & -- \\
weakNAS-t                 & jigsaw (50)          & 15.43 & 35.73 & -- & -- & -- & -- \\
Arch-Graph-zero           & jigsaw (50)          & 7.83  & 14.70 & -- & -- & -- & -- \\
Arch-Graph~\cite{huang2022archgraph} & jigsaw (50) + Task2Vec & 5.24 & \best{12.20} & 2.75 & 10.25 & -- & -- \\
TRP~\cite{li2025trp}      & zero-cost proxies    & --    & --    & 2.25 & \best{7.50} & -- & -- \\
\midrule
\multicolumn{8}{l}{\emph{Our implementation, identical protocol}}\\
Random                    & none                 & 68.36 & 87.02 & 42.25 & 55.00 & 232.0 & 294.3 \\
ZC-only ranker (TRP-style)& zero-cost proxies    & 5.49 & 25.18 & \best{1.75} & 13.50 & 34.8 & 152.1 \\
Sources-only ranker       & 6 sources $\times$ 50 & 14.26 & 17.82 & 2.50 & 12.50 & 120.2 & \best{80.5} \\
\method (jigsaw only)     & jigsaw (50) + ZC     & 4.78 & 21.27 & 2.50 & 10.75 & 27.4 & 121.8 \\
\method                   & 6 sources $\times$ 50 + ZC & \best{4.21} & 17.84 & 2.00 & 9.00 & \best{20.9} & 88.9 \\
\bottomrule
\end{tabular}
\end{table*}

\cref{tab:main} summarises the results.
Against published methods, \method obtains a better rank of the mean value than Arch-Graph on the four metric-compatible targets in both spaces (Macro 2.00 \vs 2.75, Micro 9.00 \vs 10.25) and is on par with TRP on Macro (2.00 \vs 2.25).
Under the six-target mean-rank protocol it improves over Arch-Graph on Macro (4.21 \vs 5.24) but is clearly worse on Micro (17.84 \vs 12.20); this average is dominated by room layout, whose metric differs from the published tables (we use validation loss, \cite{huang2022archgraph} reports L2 error).
It also trails TRP on Micro (9.00 \vs 7.50).
Published numbers were obtained with different code, seeds and implementation details, so these comparisons are indicative rather than controlled.
Restricted to 50 evaluated jigsaw architectures---Arch-Graph's source budget, although with zero-cost proxies instead of Task2Vec---\method is slightly better than Arch-Graph on Macro (4.78 \vs 5.24; rank of mean 2.50 \vs 2.75) but worse on Micro (rank of mean 10.75 \vs 10.25; six-target average 21.27 \vs 12.20).

Against our own ZC-only ranker the picture is more nuanced.
In the typical case the two are on par: their per-target median ranks average 3.3 \vs 3.6 on Macro and 14.8 \vs 10.5 on Micro, \method is better in mean rank on only 2 and 4 of the six targets, and the differences are not significant at either the seed or the task level.
The difference lies in the tail: the worst 10\% of ZC-only runs average rank 34.8 on Macro and 152.1 on Micro, \vs 20.9 and 88.9 for \method, and its single worst run on Micro ranks 545th of 4{,}096 (\method: 185th).
The reduction is stable across tail levels (CVaR at 5/10/20\%: $-46/{-40}/{-30}\%$ on Macro, $-48/{-42}/{-38}\%$ on Micro), but its sources differ: on Micro it appears on four of six targets and survives leaving out any single target (worst-10\% 30.4--98.6 \vs 102.3--172.8), whereas on Macro it comes entirely from room layout, without which the two methods are on par (worst-10\% 6.9 \vs 6.5).
A hierarchical bootstrap over targets and seeds puts 78\% (Macro) and 90\% (Micro) of its mass on a lighter tail for \method, but its 95\% intervals include zero; per-target numbers are in \cref{tab:tail} (supplement); the pooled analysis in \cref{sec:real} adds seven real-training targets to increase task-level power.
Sources-only fails catastrophically on Macro instead (worst 10\%: 120.2, worst run: 778th).
Combining both knowledge sources substantially mitigates the severe failures of either.

\paragraph{Proxy leakage control.}
With the original object-classification proxies, which see 32 target labels when object classification is the target (as in TRP), the results barely change: six-target means 4.30/17.70 (Macro/Micro) for \method \vs 5.42/25.21 for ZC-only, rank of mean 2.25/9.00 \vs 1.75/14.00.
Excluding object classification altogether gives 4.43/19.53 \vs 5.99/26.95.

\subsection{Complementarity of histories and proxies}
\label{sec:complementarity}

\begin{table}[t]
\centering
\small
\setlength{\tabcolsep}{4pt}
\caption{\textbf{Per-target mean rank} of a ranker pretrained on zero-cost proxies only (ZC), on sparse sources only (Src), and on both (\method). 20 seeds.}
\label{tab:complement}
\begin{tabular}{lcccccc}
\toprule
& \multicolumn{3}{c}{Macro} & \multicolumn{3}{c}{Micro}\\
\cmidrule(lr){2-4}\cmidrule(lr){5-7}
Target & ZC & Src & Both & ZC & Src & Both\\
\midrule
Object cls.   & 3.0  & 4.2  & 3.1  & 16.3 & 9.9  & 9.4\\
Scene cls.    & 1.8  & 1.7  & 2.2  & 20.2 & 5.1  & 3.6\\
Autoencoding  & 3.2  & 2.2  & 2.0  & 43.2 & 34.7 & 22.3\\
Normals       & 1.0  & 1.4  & 1.4  & 10.2 & 10.1 & 12.2\\
Sem. seg.     & 1.1  & 1.6  & 1.2  & 4.0  & 9.4  & 5.0\\
Room layout   & 23.0 & 74.4 & 15.5 & 57.2 & 37.8 & 54.6\\
\midrule
Mean          & 5.49 & 14.26& 4.21 & 25.18& 17.82& 17.84\\
\bottomrule
\end{tabular}
\end{table}

\cref{tab:complement} breaks the results down by target.
Zero-cost proxies alone are excellent on Macro, where the coarse skeleton choices that proxies capture (depth, width, down-sampling positions) dominate performance.
On Micro they fail on several targets, \eg autoencoding (43.2) and scene classification (20.2); sparse source logs correct exactly these targets.
Conversely, on Macro room layout---whose ranking is negatively correlated with the other tasks---sources alone are misleading (74.4), yet the combination is the best of the three (15.5), indicating that the shared ranker does not simply average the two signals.
In a development run we also computed the proxies on the target task instead of object classification; this degraded the ZC-only ranker on Micro from 25.2 to 36.6 (autoencoding: 93.2), confirming that proxies are unreliable on dense-prediction targets and that historical logs are needed there.

\subsection{Ablations}
\label{sec:ablation}

\begin{table}[t]
\centering
\small
\setlength{\tabcolsep}{3.5pt}
\caption{\textbf{Ablations} ($B{=}50$, $m{=}50$, 20 seeds). Mean rank over six targets, in parentheses the number of targets (of six) on which the full method is better after averaging over seeds, and mean rank of the worst 10\% of runs.}
\label{tab:ablation}
\resizebox{\linewidth}{!}{
\begin{tabular}{lcccc}
\toprule
& \multicolumn{2}{c}{Mean rank (targets better)} & \multicolumn{2}{c}{Worst 10\%}\\
\cmidrule(lr){2-3}\cmidrule(lr){4-5}
Variant & Macro & Micro & Macro & Micro\\
\midrule
\method (full) & \best{4.21} & 17.84 & \best{20.9} & 88.9\\
\quad w/o source logs & 5.49 (2/6) & 25.18 (4/6) & 34.8 & 152.1\\
\quad w/o proxies & 14.26 (4/6) & 17.82 (4/6) & 120.2 & 80.5\\
\quad w/o portfolio init. & 6.07 (6/6) & 19.64 (4/6) & 31.5 & 102.1\\
\quad w/o pretraining & 8.57 (6/6) & \best{14.07} (1/6) & 56.8 & \best{66.8}\\
\quad trees instead of ranker & 12.85 (6/6) & 16.74 (4/6) & 82.6 & 69.3\\
\quad + trees ensemble & 6.09 (6/6) & 16.47 (2/6) & 35.2 & 76.1\\
\bottomrule
\end{tabular}}
\end{table}

\cref{tab:ablation} isolates the components of \method.
On Macro, removing ranker pretraining or the portfolio worsens all six targets; with only six targets, however, no ablation remains significant after Holm correction at the task level, so we read these results as consistent directions rather than established effects.
Removing either knowledge source sharply worsens the worst-case runs.
On Micro, no component has a consistent effect across targets; notably, a ranker that uses the same features and portfolio but no pretraining is better on average (14.07, better on four of six targets), suggesting that on cell spaces the pretrained prior can be too rigid for some targets. From a risk perspective, however, no variant has a lighter tail than \method on both spaces: every variant that improves the Micro tail increases the Macro worst-10\% by at least 68\%, and we keep the configuration frozen before the final runs rather than select one per space.
Deciding adaptively when to trust pretraining is a natural extension; an exploratory online gate (\cref{sec:supp_gate}) did not reliably separate the two cases with only 50 evaluations.

\subsection{Target budget and source density}
\label{sec:sensitivity}

\begin{figure*}[t]
\centering
\includegraphics[width=0.92\linewidth]{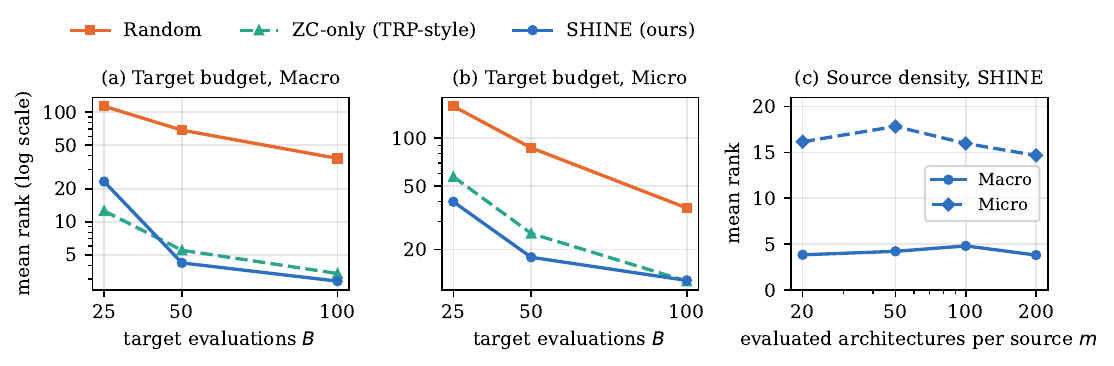}
\caption{\textbf{(a,b)} Mean rank over six targets \vs the number of target evaluations $B$ (10 seeds for $B\in\{25,100\}$, 20 for $B{=}50$). \textbf{(c)} Mean rank of \method \vs the number of evaluated architectures per source $m$ ($B{=}50$).}
\label{fig:budget}
\end{figure*}

\cref{fig:budget}(a,b) varies the target budget.
On Micro, the gap between \method and ZC-only is largest when evaluations are scarce: at $B{=}25$ the six-target mean rank is 39.9 \vs 57.3, although \method is better on only three of six targets.
On Macro, ZC-only is better at $B{=}25$ (12.6 \vs 23.2), mostly because of room layout (49.5 \vs 108.7); on the four metric-matched targets the two are on par (5.8 \vs 5.6).
At $B{=}100$ both methods find near-optimal architectures (Macro 2.9 \vs 3.4, Micro 12.8 \vs 12.6).
\cref{fig:budget}(c) varies the source density: \method is robust to $m$ between 20 and 200, with the best Micro result at $m{=}200$ (14.7) and no consistent trend on Macro.

\paragraph{Realistic search logs.}
So far each source revealed a uniformly random subset of architectures.
Real histories are biased towards good architectures, because they are the trajectories of earlier searches.
We therefore replaced each source log by the 50 architectures evaluated by a predictor-guided search (Extra-Trees weak predictor, same search template) on that source; for one seed on Macro, these logs lie on average at the 65th--85th percentile of their task (full results in \cref{tab:history}, supplement).
\method is robust to this shift: its mean rank changes from 4.21 to 3.78 on Macro and from 17.84 to 20.63 on Micro, neither difference being significant even at the seed level ($p{=}0.56$ and $0.40$), and it remains better than the ZC-only ranker in mean rank and worst-10\% on both spaces (Micro worst-10\%: 110.8 \vs 152.1).
Biased logs mostly help the sources-only ranker on Macro (14.26 $\to$ 3.64, worst-10\% 120.2 $\to$ 18.0), because past searches usually contain each source's champions; with such logs, histories alone become as strong as \method on Macro.
On Micro, the rank of the mean value on the four matched targets rises to 11.00, slightly behind Arch-Graph (10.25).

\subsection{Independent validation on MedMNIST-NAS-Bench}
\label{sec:medmnist}

\begin{table}[t]
\centering
\small
\setlength{\tabcolsep}{3.5pt}
\caption{\textbf{MedMNIST-NAS-Bench} (never used during development). Normalised test regret (lower is better) and test rank of the architecture selected by validation AUROC, averaged over four tasks and 20 seeds. No zero-cost proxies are used here.}
\label{tab:medmnist}
\resizebox{\linewidth}{!}{
\begin{tabular}{lcccc}
\toprule
& \multicolumn{2}{c}{Tiny (768, $B{=}30$)} & \multicolumn{2}{c}{Small (5{,}016, $B{=}50$)}\\
\cmidrule(lr){2-3}\cmidrule(lr){4-5}
Method & Regret & Rank & Regret & Rank\\
\midrule
Random & 0.272 & 195.1 & 0.253 & 1256 \\
Target-only ranker & 0.221 & 147.0 & 0.265 & 1317 \\
Trees + sources & 0.224 & 161.4 & \best{0.229} & 1122 \\
\method w/o ZC, $m{=}50$ & \best{0.189} & \best{115.7} & 0.230 & 1107 \\
\method w/o ZC, $m{=}200$ & 0.202 & 143.6 & \best{0.174} & \best{613} \\
\bottomrule
\end{tabular}}
\end{table}

MedMNIST-NAS-Bench provides no zero-cost proxies, so it isolates the value of sparse search histories on a search space and task family never used during development (\cref{tab:medmnist}).
On Tiny, pretraining on three sparse sources with $m{=}50$ improves over the target-only ranker on three of four tasks (seed-level $p{=}0.03$), over random search on all four, and over the tree-based transfer baseline on three of four.
On Small, histories with $m{=}50$ improve over the target-only ranker on three of four tasks and match the tree baseline on average, and denser histories ($m{=}200$) reduce the test regret by a further 24\% and nearly halve the test rank.
Because selection is based on validation AUROC, test regret also reflects the validation--test gap, which is large for the small BreastMNIST and PneumoniaMNIST datasets.

\subsection{Validation with real training and pooled analysis}
\label{sec:real}

\begin{table}[t]
\centering
\small
\setlength{\tabcolsep}{2.5pt}
\caption{\textbf{Pooled analysis over 13 Micro targets} (analysis plan fixed before the last four targets were run). Rank of the returned architecture as a percentage of the pool (lower is better): mean and CVaR$_{10}$ (mean of the worst 10\% of runs). TN: TransNAS-Bench-101 ($B{=}50$, 20 seeds, pool 4{,}096). RT: real training (pool 300, averaged over $B{\in}\{30,50\}$, 30 seeds each). Bottom: number of targets on which \method is better, and task-level Wilcoxon $p$-values (Holm-corrected over the two comparisons). $^\dagger$Leakage-free proxies (computed on scene classification).}
\label{tab:pooled}
\resizebox{\linewidth}{!}{
\begin{tabular}{lcccccc}
\toprule
& \multicolumn{3}{c}{Mean (\%)} & \multicolumn{3}{c}{CVaR$_{10}$ (\%)}\\
\cmidrule(lr){2-4}\cmidrule(lr){5-7}
Target & ZC only & Hist.\ only & \method & ZC only & Hist.\ only & \method\\
\midrule
TN object cls.$^\dagger$ & 0.40 & 0.24 & 0.23 & 1.93 & 0.68 & 0.66\\
TN scene cls.   & 0.49 & 0.12 & 0.09 & 1.42 & 0.54 & 0.38\\
TN autoencoding & 1.05 & 0.85 & 0.54 & 7.09 & 2.64 & 1.05\\
TN normals      & 0.25 & 0.25 & 0.30 & 0.59 & 0.52 & 0.77\\
TN sem.\ seg.   & 0.10 & 0.23 & 0.12 & 0.20 & 0.93 & 0.24\\
TN room layout  & 1.40 & 0.92 & 1.33 & 7.48 & 3.67 & 4.50\\
RT CIFAR-100    & 1.02 & 0.71 & 0.60 & 3.06 & 3.56 & 2.50\\
RT Tiny-IN cls. & 0.42 & 0.81 & 0.60 & 1.22 & 4.67 & 2.56\\
RT Pets cls.    & 10.39 & 9.51 & 8.02 & 39.00 & 33.17 & 35.50\\
RT Tiny-IN rec. & 1.63 & 0.69 & 0.98 & 6.50 & 3.00 & 4.50\\
RT denoising    & 1.92 & 0.71 & 1.05 & 7.78 & 3.44 & 4.33\\
RT inpainting   & 1.74 & 0.66 & 1.24 & 6.22 & 2.28 & 4.17\\
RT Pets seg.    & 2.07 & 1.29 & 1.69 & 6.56 & 1.44 & 4.50\\
\midrule
Mean            & 1.76 & 1.31 & 1.29 & 6.85 & 4.66 & 5.05\\
\method better  & 10/13 & 7/13 & -- & 10/13 & 6/13 & --\\
Holm $p$        & .012 & .68 & -- & .016 & .45 & --\\
\bottomrule
\end{tabular}}
\end{table}

To test whether the conclusions carry over beyond tabular benchmarks, we trained 300 randomly drawn architectures from each TransNAS-Bench-101 space with the official model code on seven targets absent from Taskonomy, each for 20 epochs with one seed: classification on CIFAR-100, Tiny-ImageNet and Oxford-IIIT Pets (37 breeds), and four dense-prediction targets---reconstruction, denoising and inpainting on Tiny-ImageNet and trimap segmentation on Pets---for which the encoder is followed by the official TransNAS autoencoder decoder with channel widths divided by four.
All inputs are $64{\times}64$; we search within these pools using the TransNAS tasks as sparse sources ($m{=}50$) with the configuration frozen in \cref{sec:exp}, at $B{\in}\{30,50\}$.
We also trained both pools on SVHN. Like other decade-old digit benchmarks, it is saturated for modern CNNs: all architectures reach about 96\% test accuracy and every method selects an architecture within 0.3 points of the pool optimum, so test-rank differences are dominated by noise (the ZC-only ranker happened to rank best on Micro); we therefore replaced it by harder targets.
Each training protocol was fixed before its targets were trained. The pooled analysis plan was fixed (internal record, 2026-10-22) after the six TransNAS targets and the CIFAR-100 and Tiny-ImageNet classification and reconstruction results had been seen, and before the four remaining targets (denoising, inpainting, Pets classification and segmentation) were run; we therefore report the new targets separately. Per-target results are in \cref{tab:real} (supplement).

On the Macro pools all transfer methods return architectures within a few ranks of the pool optimum (mean rank at most 5.3 of 300), so these pools hardly discriminate between methods.
We therefore pool the six TransNAS Micro targets and the seven real-training Micro targets (\cref{tab:pooled}).
Proxies fail on the dense-prediction targets, where histories are the best single source, and they lose to \method on 10 of 13 targets both in mean and in worst-case performance (task-level Holm-corrected $p{=}0.012$ and $0.016$; bootstrap interval of the mean CVaR$_{10}$ difference $[-2.8,-0.8]$ percentage points).
On the four targets run after the plan was fixed, \method beats the proxy-only ranker on all four in both statistics (the smallest attainable $p$ with four targets is 0.125), but histories alone have the lighter tail on all four.
The advantage over proxies does not hinge on a single data source: it holds on the seven real-training targets alone (6/7, $p{=}0.031$ and $0.047$), without the four Tiny-ImageNet targets (7/9, $p{=}0.020$), and when targets sharing a real dataset are averaged into one unit (7/9, $p{=}0.020$); details are in \cref{tab:sens} (supplement). Since the pools differ in size and budget fraction, the grouped results should also be read separately: on the six TransNAS targets \method beats proxies on 4/6 in both statistics.
Histories alone are on par with \method on these cell-space targets (7/13 and 6/13, $p{=}0.68$ and $0.45$) and slightly lighter-tailed on average, but they fail severely on Macro (\cref{tab:main}: worst-10\% rank 120.2 \vs 20.9). Thus \method does not dominate either source; it substantially mitigates the severe failures observed with each of them while retaining competitive average performance.

\section{Limitations and Conclusion}
\label{sec:conclusion}

\paragraph{Limitations.}
Our main experiments use tabular benchmarks, which allow exhaustive evaluation and controlled leakage-free analyses; our real-training validation (\cref{sec:real}) is limited to pools of 300 architectures, short schedules, a single training seed and $64{\times}64$ inputs, and its Macro pools do not discriminate between methods.
The number of targets is small: six per TransNAS space, and the 13 pooled targets share four underlying datasets, so task-level tests have limited power and are not fully independent.
Hyper-parameters were developed on TransNAS-Bench-101; we therefore validate on MedMNIST-NAS-Bench, which was never used during development. On Micro, \method still trails the reported results of TRP.
Finally, \method assumes that past tasks share the search space of the new task; transferring across different search spaces is left to future work.

\paragraph{Conclusion.}
We studied cross-task NAS from sparse search histories, the form in which search knowledge actually accumulates.
Consensus priors and source portfolios that work with dense tables fail in this regime, whereas zero-cost proxies, though available everywhere, are unreliable on dense-prediction tasks.
\method unifies both kinds of cheap knowledge by pretraining a pairwise ranker on sparse source tasks and proxy pseudo-tasks and by initialising the search with a champion portfolio.
\method is best understood as a risk-robust strategy: it substantially mitigates the severe failures that each knowledge source has on some target, reduces tail risk relative to proxies, while retaining competitive average performance. Its advantage over proxies alone is small on the six TransNAS targets but significant at the task level when pooled over 13 cell-space targets, whereas histories alone are on par with it on cell spaces but fail severely on the macro space.
Deciding adaptively how much to trust each source of prior knowledge, which could also recover the typical-case gains of simpler variants on cell spaces, is a promising direction.
We hope that framing transfer around incomplete histories will make cross-task NAS easier to deploy where past searches are already available.

\section*{Acknowledgments}
This work was supported in part by JST SPRING, Grant Number JPMJSP2121 and JSPS Program for Forming Japan's Peak Research Universities (J-PEAKS), Grant Number JPJS00420240016.
{

    \small
    \bibliographystyle{ieeenat_fullname}
    \bibliography{main}
}

\clearpage
\setcounter{page}{1}
\maketitlesupplementary
\appendix

\section{Implementation details}
\label{sec:supp_impl}

\paragraph{Encodings.}
For the Micro cell (four nodes, six edges with operations \{zero, skip, $1{\times}1$ conv, $3{\times}3$ conv\}) we use one-hot edge operations plus a BANANAS-style path encoding~\cite{white2021bananas}: each of the four input-to-output paths is active if it contains no zero operation and is encoded by the one-hot index of its operation sequence ($3+9+9+27$ features), complemented by the number of active paths, the number of $3{\times}3$ convolutions on active paths and a parameter-count surrogate.
For Macro (four to six layers of types 1--4, where even types double the width and types $>2$ halve the resolution) we use per-layer one-hot types, per-layer down-sampling and widening indicators, their cumulative sums, depth, and the mean positions of down-sampling and widening layers.
All features are standardised.

\paragraph{Zero-cost proxies.}
We use the twelve proxies of NAS-Bench-Suite-Zero~\cite{krishnakumar2022nbsz} (fisher, flops, grad-norm, grasp, jacov, l2-norm, nwot, params, plain, snip, synflow, zen) computed on object classification, rank-normalised to $[0,1]$.

\paragraph{Imputation.}
Each sparse source is imputed by ridge regression ($\alpha{=}10$) on the first-order encoding and all pairwise products, fitted to the percentile ranks of its $m$ observed architectures.

\paragraph{Compute.}
One full search ($B{=}50$) takes about 77\,s on an RTX~5090 shared by 12 processes; the ZC-only ranker takes 54\,s and the tree-based variant under 1\,s.

\section{Exploratory: an online gate between pretrained and target-only rankers}
\label{sec:supp_gate}

The ablation in \cref{tab:ablation} suggests that pretraining helps on Macro but can be too rigid on Micro.
\emph{After} observing the final results, we therefore tested a prequential gate and fixed its rule and pass criteria before running it.
At every round we fit both the pretrained ranker (A) and a ranker with the same features and initialisation but without pretraining (B).
After each new batch is evaluated, the predictions that each model made \emph{before} observing the batch are scored by pairwise ordering accuracy on (new, old) and (new, new) pairs; with accumulated correct counts $C_A, C_B$ over $P$ pairs, the weight is $w=\sigma\big(2\,(C_A-C_B)/\sqrt{P}\big)$ and the search score is $w\,z(A)+(1-w)\,z(B)$.
A fixed $w{=}0.5$ ensemble served as control.

\begin{table}[h]
\centering
\small
\setlength{\tabcolsep}{4pt}
\caption{\textbf{Exploratory gate} (20 seeds; run with the original object-classification proxies). TransNAS: mean rank over six targets / rank of mean value on the four matched targets. MedMNIST: test regret ($m{=}50$, no proxies). The gate did not meet its pre-declared criterion (within 1 rank, resp.\ 0.01 regret, of the better single model).}
\label{tab:gate}
\begin{tabular}{lcccc}
\toprule
& \multicolumn{2}{c}{TransNAS} & \multicolumn{2}{c}{MedMNIST}\\
\cmidrule(lr){2-3}\cmidrule(lr){4-5}
Method & Macro & Micro & Tiny & Small\\
\midrule
Pretrained (\method) & 4.30 / 2.25 & 17.70 / 9.00 & 0.189 & 0.230\\
No pretraining & 9.08 / 2.75 & 14.45 / 9.00 & 0.221 & 0.265\\
Online gate & 5.32 / 2.75 & 15.52 / 9.50 & 0.215 & 0.232\\
Fixed 0.5 ensemble & 5.13 / 2.00 & 17.29 / 8.00 & 0.223 & 0.224\\
\bottomrule
\end{tabular}
\end{table}

The gate missed its criterion on both TransNAS spaces (by 0.02 and 0.07 ranks) and on MedMNIST Tiny.
Its final weights stayed between 0.4 and 0.6 on average for every target, \ie with 50 noisy evaluations the out-of-sample evidence was insufficient to separate the two rankers.
The fixed ensemble obtained the best metric-matched TransNAS numbers (2.00 / 8.00), but since it was not a pre-declared candidate we do not adopt it and report it only for completeness.
Reliably deciding \emph{when} to trust transferred knowledge under very small budgets remains an open problem.

\section{Search-trajectory source logs}
\label{sec:supp_history}

\begin{table}[h]
\centering
\small
\setlength{\tabcolsep}{3.5pt}
\caption{\textbf{Uniform \vs search-trajectory source logs} ($B{=}50$, $m{=}50$, 20 seeds). Trajectory logs are the 50 architectures evaluated by an Extra-Trees weak-predictor search on each source. Mean rank over six targets / rank of the mean value on the four matched targets / worst-10\%.}
\label{tab:history}
\resizebox{\linewidth}{!}{
\begin{tabular}{llcc}
\toprule
Method & Logs & Macro & Micro\\
\midrule
ZC-only ranker & -- & 5.49 / 1.75 / 34.8 & 25.18 / 13.50 / 152.1\\
Sources-only ranker & uniform & 14.26 / 2.50 / 120.2 & 17.82 / 12.50 / 80.5\\
Sources-only ranker & trajectory & 3.64 / 2.00 / 18.0 & 18.74 / 9.00 / 92.4\\
\method & uniform & 4.21 / 2.00 / 20.9 & 17.84 / 9.00 / 88.9\\
\method & trajectory & 3.78 / 2.00 / 17.8 & 20.63 / 11.00 / 110.8\\
\bottomrule
\end{tabular}}
\end{table}

Paired over (target, seed), \method with trajectory logs does not differ significantly from \method with uniform logs ($p{=}0.56$ on Macro, $0.40$ on Micro) or from the sources-only ranker with trajectory logs ($p{=}0.81$ and $0.30$).
The protocol of this experiment was fixed before it was run.

\section{Tail robustness per target}
\label{sec:supp_tail}

\begin{table}[h]
\centering
\small
\setlength{\tabcolsep}{3pt}
\caption{\textbf{Worst-10\% rank per target} (mean of the two worst of 20 seeds; $B{=}50$, $m{=}50$) for \method, the ZC-only ranker and the sources-only ranker, and pooled CVaR over all 120 runs with the 95\% hierarchical bootstrap interval (targets, then seeds) of the difference \method$-$ZC.}
\label{tab:tail}
\resizebox{\linewidth}{!}{
\begin{tabular}{lcccccc}
\toprule
& \multicolumn{3}{c}{Macro} & \multicolumn{3}{c}{Micro}\\
\cmidrule(lr){2-4}\cmidrule(lr){5-7}
Target & \method & ZC & Src & \method & ZC & Src\\
\midrule
Object cls. & 9.0 & 6.0 & 13.0 & 27.0 & 79.0 & 28.0\\
Scene cls. & 11.0 & 8.5 & 5.5 & 15.5 & 58.0 & 22.0\\
Autoencoding & 5.0 & 5.5 & 9.0 & 43.0 & 290.5 & 108.0\\
Normals & 3.0 & 1.0 & 4.0 & 31.5 & 24.0 & 21.5\\
Sem.\ seg. & 3.0 & 2.0 & 4.0 & 10.0 & 8.0 & 38.0\\
Room layout & 44.5 & 73.5 & 465.0 & 184.5 & 306.5 & 150.5\\
\midrule
CVaR 5\% & 28.8 & 53.5 & 208.0 & 128.8 & 248.3 & 111.2\\
CVaR 10\% & 20.9 & 34.8 & 120.2 & 88.9 & 152.1 & 80.5\\
CVaR 20\% & 14.8 & 21.0 & 64.8 & 58.2 & 93.4 & 54.8\\
\midrule
\multicolumn{7}{l}{95\% CI of \method$-$ZC: CVaR 10\%: Macro $[-38.2, 3.2]$, Micro $[-217.1, 25.1]$}\\
\bottomrule
\end{tabular}}
\end{table}

\section{Per-target real-training results}
\label{sec:supp_real}

\begin{table}[h]
\centering
\small
\setlength{\tabcolsep}{2.5pt}
\caption{\textbf{Real training, per target.} Test rank (lower is better) of the architecture selected on the validation set within a pool of 300 architectures, mean over 30 seeds, worst-10\% in parentheses. Metrics: accuracy (classification), SSIM (reconstruction, denoising), PSNR inside the hole (inpainting), mIoU (segmentation).}
\label{tab:real}
\resizebox{\linewidth}{!}{
\begin{tabular}{llccccc}
\toprule
Target & $B$ & Random & Target-only & ZC only & Hist.\ only & \method\\
\midrule
\multicolumn{7}{l}{\emph{Macro}}\\
CIFAR-100 & 30 & 11.0 (29) & 1.1 (2) & 1.0 (1) & 1.0 (1) & 1.0 (1)\\
 & 50 & 7.8 (28) & 1.0 (1) & 1.0 (1) & 1.0 (1) & 1.0 (1)\\
Tiny-IN cls. & 30 & 11.0 (26) & 2.0 (2) & 2.0 (2) & 2.0 (2) & 2.0 (2)\\
 & 50 & 6.7 (25) & 2.0 (2) & 2.0 (2) & 2.0 (2) & 2.0 (2)\\
Pets cls. & 30 & 10.6 (37) & 3.1 (17) & 5.3 (14) & 1.5 (6) & 2.4 (14)\\
 & 50 & 10.0 (31) & 1.0 (1) & 1.4 (5) & 1.0 (1) & 1.0 (1)\\
Tiny-IN rec. & 30 & 8.6 (22) & 1.7 (5) & 1.0 (1) & 1.0 (1) & 1.0 (1)\\
 & 50 & 7.0 (20) & 1.0 (1) & 1.0 (1) & 1.0 (1) & 1.0 (1)\\
Denoising & 30 & 10.7 (37) & 2.9 (8) & 2.0 (8) & 2.3 (6) & 1.7 (6)\\
 & 50 & 4.4 (21) & 1.2 (2) & 1.0 (1) & 1.0 (1) & 1.0 (1)\\
Inpainting & 30 & 11.0 (35) & 1.8 (5) & 1.0 (1) & 1.0 (1) & 1.0 (1)\\
 & 50 & 6.8 (20) & 1.0 (1) & 1.0 (1) & 1.0 (1) & 1.0 (1)\\
Pets seg. & 30 & 9.9 (35) & 1.1 (2) & 1.8 (4) & 1.0 (1) & 1.0 (1)\\
 & 50 & 6.7 (24) & 1.0 (1) & 1.0 (1) & 1.0 (1) & 1.0 (1)\\
\midrule\multicolumn{7}{l}{\emph{Micro}}\\
CIFAR-100 & 30 & 16.1 (44) & 6.8 (30) & 4.4 (11) & 3.0 (17) & 2.5 (10)\\
 & 50 & 10.4 (37) & 1.5 (5) & 1.8 (5) & 1.2 (3) & 1.1 (2)\\
Tiny-IN cls. & 30 & 33.7 (102) & 5.9 (33) & 1.5 (6) & 3.7 (21) & 2.1 (8)\\
 & 50 & 14.9 (52) & 2.3 (9) & 1.0 (1) & 1.2 (3) & 1.5 (6)\\
Pets cls. & 30 & 51.5 (117) & 41.3 (117) & 34.7 (117) & 34.0 (96) & 32.3 (117)\\
 & 50 & 27.4 (88) & 27.2 (96) & 27.7 (110) & 23.1 (103) & 15.8 (46)\\
Tiny-IN rec. & 30 & 6.8 (24) & 5.7 (26) & 8.3 (23) & 2.8 (13) & 4.7 (14)\\
 & 50 & 4.6 (16) & 1.3 (3) & 1.4 (5) & 1.3 (4) & 1.1 (2)\\
Denoising & 30 & 9.1 (34) & 5.5 (15) & 8.4 (31) & 3.3 (12) & 5.0 (14)\\
 & 50 & 5.4 (20) & 2.2 (9) & 3.1 (13) & 1.0 (1) & 1.3 (4)\\
Inpainting & 30 & 8.5 (32) & 5.5 (30) & 8.2 (22) & 2.5 (9) & 5.4 (13)\\
 & 50 & 4.7 (14) & 2.1 (8) & 2.3 (10) & 1.4 (5) & 2.1 (5)\\
Pets seg. & 30 & 9.2 (30) & 4.3 (10) & 6.7 (24) & 3.8 (5) & 6.2 (15)\\
 & 50 & 5.6 (18) & 4.1 (12) & 5.7 (15) & 4.0 (4) & 3.9 (5)\\
\bottomrule
\end{tabular}}
\end{table}

\section{Sensitivity of the pooled analysis}
\label{sec:supp_sens}

\begin{table}[h]
\centering
\small
\setlength{\tabcolsep}{3pt}
\caption{\textbf{Sensitivity of the pooled comparison} (not part of the fixed plan). Number of targets on which \method is better and task-level Wilcoxon $p$ (uncorrected), for mean rank and CVaR$_{10}$ (\% of pool).}
\label{tab:sens}
\resizebox{\linewidth}{!}{
\begin{tabular}{lccccc}
\toprule
& & \multicolumn{2}{c}{vs ZC only} & \multicolumn{2}{c}{vs Hist.\ only}\\
\cmidrule(lr){3-4}\cmidrule(lr){5-6}
Subset & $n$ & Mean & CVaR$_{10}$ & Mean & CVaR$_{10}$\\
\midrule
All targets & 13 & 10 (.006) & 10 (.008) & 7 (.69) & 6 (.46)\\
Seen before the plan & 9 & 6 (.10) & 6 (.13) & 6 (.73) & 6 (.43)\\
Run after the plan & 4 & 4 (.13) & 4 (.13) & 1 (.88) & 0 (.13)\\
TransNAS only & 6 & 4 (.16) & 4 (.16) & 4 (.84) & 4 (.69)\\
Real training only & 7 & 6 (.031) & 6 (.047) & 3 (.58) & 2 (.30)\\
Without Tiny-ImageNet & 9 & 7 (.020) & 7 (.020) & 6 (.65) & 5 (.82)\\
Without Pets & 11 & 8 (.024) & 8 (.032) & 6 (.58) & 6 (.97)\\
Real datasets as units & 9 & 7 (.020) & 7 (.020) & 6 (.57) & 5 (1.0)\\
\bottomrule
\end{tabular}}
\end{table}

\end{document}